\documentclass[10pt,twocolumn,letterpaper]{article}

\usepackage{cvpr}
\usepackage{times}
\usepackage{epsfig}
\usepackage{graphicx}
\usepackage{amsmath}
\usepackage{amssymb}
\usepackage{subcaption}
\usepackage{multirow}
\usepackage{color}
\usepackage[space]{grffile}
\usepackage{array}
\newcolumntype{C}[1]{>{\centering\arraybackslash}p{#1}}
\usepackage{chngpage}

\usepackage[pagebackref=true,breaklinks=true,letterpaper=true,colorlinks,bookmarks=false]{hyperref}
\usepackage[font=small,labelfont=bf,tableposition=top]{caption}
\cvprfinalcopy 

\def\cvprPaperID{976} 

\ifcvprfinal\fi
\begin{document}

\title{Deeply-Recursive Convolutional Network for Image Super-Resolution}

\author{Jiwon Kim, Jung Kwon Lee and Kyoung Mu Lee\\
	Department of ECE, ASRI, Seoul National University, Korea\\
	{\tt\small \{j.kim, deruci, kyoungmu\}@snu.ac.kr}
}

\maketitle


\begin{abstract}
We propose an image super-resolution method (SR) using a deeply-recursive convolutional network (DRCN). Our network has a very deep recursive layer (up to 16 recursions). Increasing recursion depth can improve performance without introducing new parameters for additional convolutions. Albeit advantages, learning a DRCN is very hard with a standard gradient descent method due to exploding/vanishing gradients. To ease the difficulty of training, we propose two extensions: recursive-supervision and skip-connection. Our method outperforms previous methods by a large margin.
\end{abstract}

\begin{figure*}[t]
	\includegraphics[width=\textwidth]{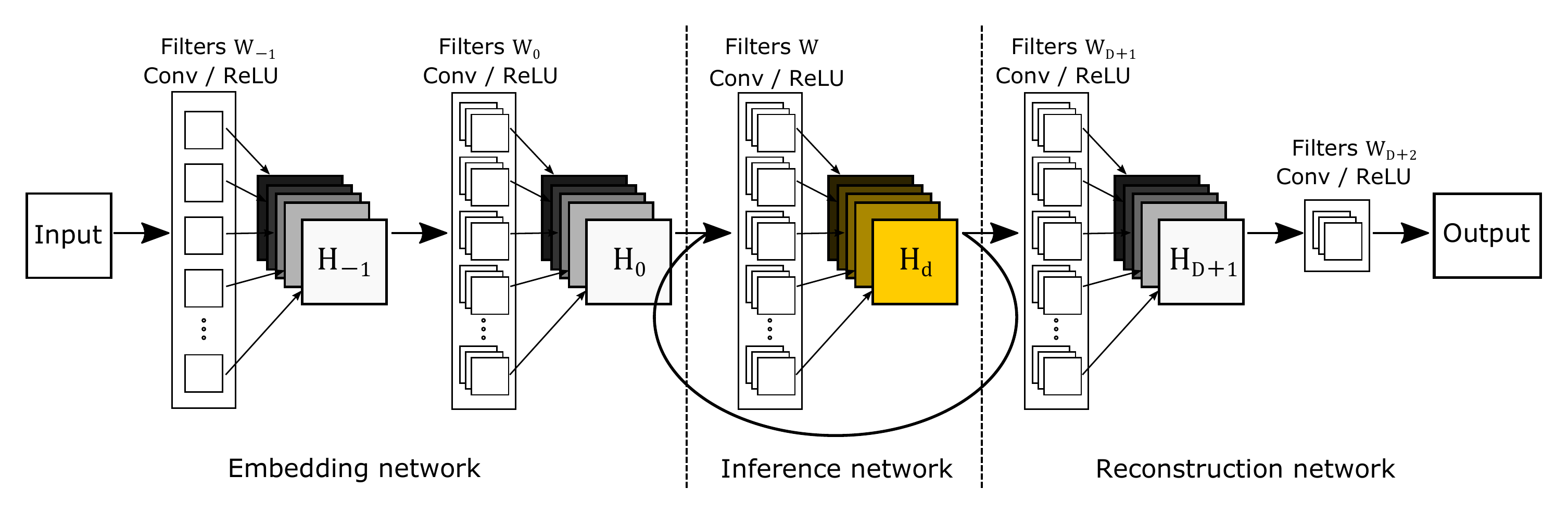}
	\caption {Architecture of our basic model. It consists of three parts: embedding network, inference network and reconstruction network. Inference network has a recursive layer and its unfolded version is in Figure \ref{fig:inference_network}.}
	\label{fig:overview}
\end{figure*}

\section{Introduction}
For image super-resolution (SR), receptive field of a convolutional network determines the amount of contextual information that can be exploited to infer missing high-frequency components. For example, if there exists a pattern with smoothed edges contained in a receptive field, it is plausible that the pattern is recognized and edges are appropriately sharpened. As SR is an ill-posed inverse problem, collecting and analyzing more neighbor pixels can possibly give more clues on what may be lost by downsampling. 

Deep convolutional networks (DCN) succeeding in various computer vision tasks often use very large receptive fields  (224x224 common in ImageNet classification \cite{krizhevsky2012imagenet, simonyan2015very}). Among many approaches to widen the receptive field, increasing network depth is one possible way: a convolutional (conv.) layer  with filter size larger than a $1\times 1$ or a pooling (pool.) layer that reduces the dimension of intermediate representation can be used.  Both approaches have drawbacks: a conv. layer introduces more parameters and a pool. layer typically discards some pixel-wise information. 

For image restoration problems such as super-resolution and denoising, image details are very important. Therefore, most deep-learning approaches for such problems do not use pooling. Increasing depth by adding a new weight layer basically introduces more parameters. Two problems can arise. First, overfitting is highly likely. More data are now required. Second, the model becomes too huge to be stored and retrieved.


To resolve these issues, we use a deeply-recursive convolutional network (DRCN). DRCN repeatedly applies the same convolutional layer as many times as desired. The number of parameters do not increase while more recursions are performed. Our network has the receptive field of 41 by 41 and this is relatively large compared to SRCNN \cite{dong2014image} (13 by 13). While DRCN has good properties, we find that DRCN optimized with the widely-used stochastic gradient descent method does not easily converge. This is due to exploding/vanishing gradients \cite{bengio1994learning}. Learning long-range dependencies between pixels with a single weight layer is very difficult. 

We propose two approaches to ease the difficulty of training (Figure \ref{fig:recursive_supervision}(a)). First, all recursions are supervised. Feature maps after each recursion are used to reconstruct the target high-resolution image (HR). Reconstruction method (layers dedicated to reconstruction) is the same for all recursions. As each recursion leads to a different HR prediction, we combine all predictions resulting from different levels of recursions to deliver a more accurate final prediction. The second proposal is to use a skip-connection from input to the reconstruction layer. In SR, a low-resolution image (input) and a high-resolution image (output) share the same information to a large extent. Exact copy of input, however, is likely to be attenuated during many forward passes. We explicitly connect the input to the layers for output reconstruction. This is particularly effective when input and output are highly correlated.

\textbf{Contributions} In summary, we propose an image super-resolution method deeply recursive in nature. It utilizes a very large context compared to previous SR methods with only a single recursive layer. We improve the simple recursive network in two ways: recursive-supervision and skip-connection. Our method demonstrates state-of-the-art performance in common benchmarks.

\section{Related Work}

\subsection{Single-Image Super-Resolution}

We apply DRCN to single-image super-resolution (SR) \cite{Irani1991, freeman2000learning,glasner2009super}. Many SR methods have been proposed in the computer vision community. Early methods use very fast interpolations but yield poor results. Some of the more powerful methods utilize statistical image priors \cite{sun2008image,Kim2010} or internal patch recurrence \cite{glasner2009super, Huang-CVPR-2015}. Recently, sophisticated learning methods have been widely used to model a mapping from LR to HR patches. Many methods have paid attention to find better regression functions from LR to HR images. This is achieved with various techniques: neighbor embedding \cite{chang2004super,bevilacqua2012}, sparse coding \cite{yang2010image,zeyde2012single,Timofte2013,Timofte}, convolutional neural network (CNN) \cite{dong2014image} and random forest \cite{schulter2015fast}.

Among several recent learning-based successes,  convolutional neural network (SRCNN) \cite{dong2014image} demonstrated the feasibility of an end-to-end approach to SR. One possibility to improve SRCNN is to simply stack more weight layers as many times as possible. However, this significantly increases the number of parameters and requires more data to prevent overfitting. In this work, we seek to design a convolutional network that models long-range pixel dependencies with limited capacity. Our network recursively widens the receptive field without increasing model capacity. 

\subsection{Recursive Neural Network in Computer Vision}

Recursive neural networks, suitable for temporal and sequential data, have seen limited use on algorithms operating on a single static image.   Socher et al.  \cite{socher2012convolutional} used a convolutional network in a separate stage to first learn features on RGB-Depth data, prior to hierarchical merging. In these models, the input dimension is twice that of the output and recursive convolutions are applied only two times. Similar dimension reduction occurs in the recurrent convolutional neural networks used for semantic segmentation \cite{pinheiro2014recurrent}. As SR methods predict full-sized images, dimension reduction is not allowed.

In Eigen et al. \cite{Eigen2014}, recursive layers have the same input and output dimension, but recursive convolutions resulted in worse performances than a single convolution due to overfitting. To overcome overfitting, Liang and Hu \cite{Liang_2015_CVPR} uses a recurrent layer that takes feed-forward inputs into all unfolded layers. They show that performance increases up to three convolutions. Their network structure, designed for object recognition, is the same as the existing CNN architectures.

Our network is similar to the above in the sense that recursive or recurrent layers are used with convolutions. We further increase the recursion depth and demonstrate that very deep recursions can significantly boost the performance for super-resolution. We apply the same convolution up to 16 times (the previous maximum is three). 

\begin{figure}[t]
	\includegraphics[width=0.5\textwidth]{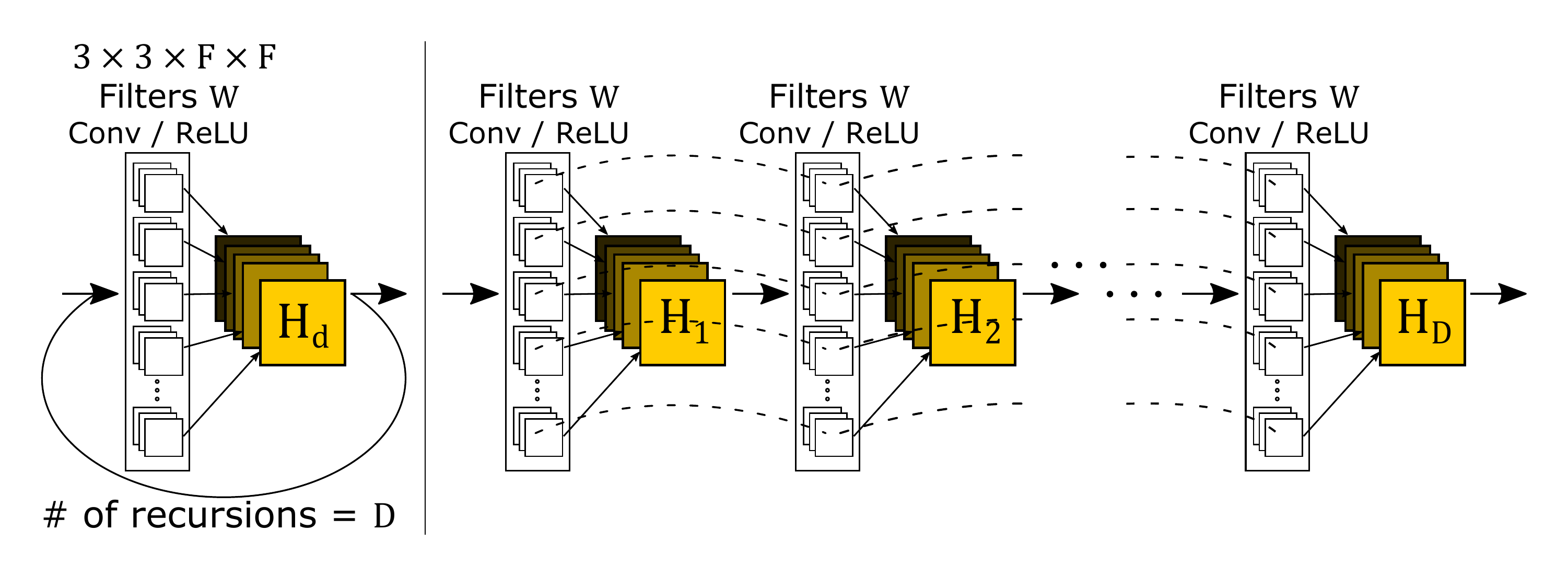}
	\caption {Unfolding inference network. \textbf{Left}: A recursive layer \textbf{Right}: Unfolded structure. The same filter W is applied to feature maps recursively. Our model can utilize very large context without adding new weight parameters. }
	\label{fig:inference_network}
\end{figure}

\begin{figure*}[t]
\begin{center}
	\includegraphics[width=\textwidth]{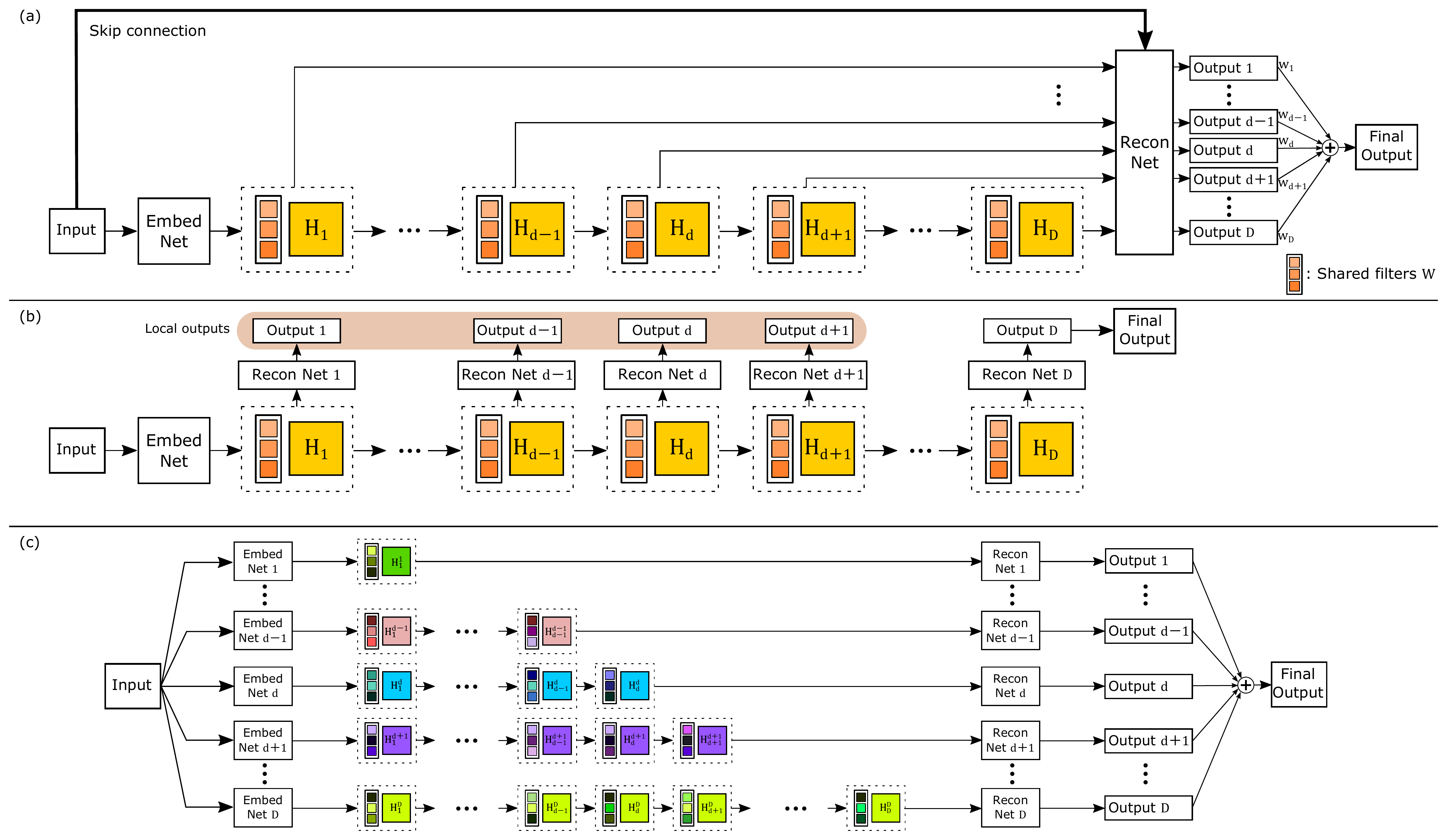}
	\caption{(a): Our final (advanced) model with recursive-supervision and skip-connection. The reconstruction network is shared for recursive predictions. We use all predictions from the intermediate recursion to obtain the final output. (b): Applying deep-supervision \cite{lee2014deeply} to our basic model. Unlike in (a), the model in (b) uses different reconstruction networks for recursions and more parameters are used.  (c): An example of expanded structure of (a) without parameter sharing (no recursion). The number of weight parameters is proportional to the depth squared. }
\label{fig:recursive_supervision}
\end{center}
\end{figure*}

\section{Proposed Method}
\subsection{Basic Model}

Our first model, outlined in Figure \ref{fig:overview}, consists of three sub-networks: embedding, inference and reconstruction networks. The embedding net is used to represent the given image as feature maps ready for inference. Next, the inference net solves the task. Once inference is done, final feature maps in the inference net are fed into the reconstruction net to generate the output image.

The \textbf{embedding net} takes the input image (grayscale or RGB) and represents it as a set of feature maps. Intermediate representation used to pass information to the inference net largely depends on how the inference net internally represent its feature maps in its hidden layers. Learning this representation is done end-to-end altogether with learning other sub-networks. \textbf{Inference net} is the main component that solves the task of super-resolution. Analyzing a large image region is done by a single recursive layer. Each recursion applies the same convolution followed by a rectified linear unit (Figure \ref{fig:inference_network}). With convolution filters larger than $1\times 1$,  the receptive field is widened with every recursion. While feature maps from the final application of the recursive layer represent the high-resolution image, transforming them (multi-channel) back into the original image space (1 or 3-channel) is necessary. This is done by the \textbf{reconstruction net}.  

We have a single hidden layer for each sub-net. Only the layer for the inference net is recursive. Other sub-nets are vastly similar to the standard mutilayer perceptrons (MLP) with a single hidden layer. For MLP, full connection of $F$ neurons is equivalent to a convolution with $1\times 1\times F \times F$. In our sub-nets, we use $3\times 3\times F \times F$ filters. For embedding net, we use $3\times 3$ filters because image gradients are more informative than the raw intensities for super-resolution. For inference net, $3\times 3$ convolutions imply that hidden states are passed to adjacent pixels only. Reconstruction net also takes direct neighbors into account.

\textbf{Mathematical Formulation} The network takes an interpolated input image (to the desired size) as input ${\bf x}$ and predicts the target image ${\bf y}$ as in SRCNN \cite{dong2014image}. Our goal is to learn a model $f$ that predicts values $\mathbf{\hat{y}}=f(\mathbf{x})$, where $\mathbf{\hat{y}}$ is its estimate of ground truth output $\mathbf{y}$. Let $f_1, f_2, f_3$ denote sub-net functions: embedding, inference and reconstruction, respectively. Our model is the composition of three functions: $f({\bf x}) = f_3(f_2 (f_1({\bf x}))).$
 
 Embedding net $f_1({\bf x})$ takes the input vector ${\bf x}$ and computes the matrix output $H_0$, which is an input to the inference net $f_2$. Hidden layer values are denoted by $H_{-1}$. The formula for embedding net is as follows:
  \begin{align}
        H_{-1} &= max(0, W_{-1}*{\bf x} + b_{-1})\\
        H_0 &= max(0, W_{0}*H_{-1} + b_0)\\
        f_1({\bf x}) &= H_0,
    \end{align}
where the operator $*$ denotes a convolution and $max(0,\cdot)$ corresponds to a ReLU. Weight and bias matrices are $W_{-1},W_0$ and $b_{-1},b_0$.

Inference net $f_2$ takes the input matrix $H_0$ and computes the matrix output $H_{D}$. Here, we use the same weight and bias matrices $W$ and $b$ for all operations.  Let $g$ denote the function modeled by a single recursion of the recursive layer: $g(H)=max(0,W*H+b)$. The recurrence relation is  
\begin{equation}
 H_d = g(H_{d-1}) = max(0,W*H_{d-1}+b),
\end{equation}
for $d = 1, ..., D$. 
Inference net $f_2$ is equivalent to the composition of the same elementary function $g$: 
\begin{equation}
f_2(H) = (g \circ g \circ \cdots \circ) g(H) =  g^{D}(H),
\end{equation}
where the operator $\circ$ denotes a function composition and $g^{d}$ denotes the $d$-fold product of $g$.

Reconstruction net $f_3$ takes the input hidden state $H_D$ and outputs the target image (high-resolution). Roughly speaking, reconstruction net is the inverse operation of embedding net. The formula is as follows:
\begin{align}
	H_{D+1} &= max(0, W_{D+1}*H_D + b_{D+1})\\
	\hat{{\bf y}} &= max(0, W_{D+2}*H_{D+1} + b_{D+2})\\
	f_3(H) &= \hat{{\bf y}}.
\end{align}

\textbf{Model Properties} Now we have all components for our model. The recursive model has pros and cons. While the recursive model is simple and powerful, we find training a deeply-recursive network very difficult. This is in accordance with the limited success of previous methods using at most three recursions so far \cite{Liang_2015_CVPR}.  Among many reasons, two severe problems are \textit{vanishing} and \textit{exploding gradients} \cite{bengio1994learning, pascanu2013difficulty}.  

\textit{Exploding gradients} refer to the large increase in the norm
of the gradient during training. Such events are due to
the multiplicative nature of chained gradients. Long term components can grow exponentially for deep recursions. The
\textit{vanishing gradients} problem refers to the opposite behavior. Long term components approach exponentially
fast to the zero vector. Due to this, learning the relation between distant pixels is very hard. Another known issue is that storing an exact copy of information through many recursions is not easy. In SR, output is vastly similar to input and recursive layer needs to keep the exact copy of input image for many recursions. These issues are also observed when we train our basic recursive model and we did not succeed in training a deeply-recursive network. 

In addition to gradient problems, there exists an issue with finding the optimal number of recursions. If recursions are too deep for a given task, we need to reduce the number of recursions. Finding the optimal number requires training many networks with different recursion depths.   

\subsection{Advanced Model} 
\textbf{Recursive-Supervision} To resolve the gradient and optimal recursion issues, we propose an improved model. We supervise all recursions in order to alleviate the effect of vanishing/exploding gradients. As we have assumed that the same representation can be used again and again during convolutions in the inference net, the same reconstruction net is used to predict HR images for all recursions. Our reconstruction net now outputs $D$ predictions and all predictions are simultaneously supervised during training (Figure \ref{fig:recursive_supervision} (a)). We use all $D$ intermediate predictions to compute the final output. All predictions are averaged during testing. The optimal weights are automatically learned during training. 

A similar but a different concept of supervising intermediate layers for a convolutional network is used in Lee et al  \cite{lee2014deeply}. Their method simultaneously minimizes classification error while improving the directness and transparency of the hidden layer learning process. There are two significant differences between our recursive-supervision and deep-supervision proposed in Lee et al. \cite{lee2014deeply}. They associate a unique classifier for each hidden layer. For each additional layer, a new classifier has to be introduced, as well as new parameters. If this approach is used, our modified network would resemble that of Figure \ref{fig:recursive_supervision}(b). We would then need $D$ different reconstruction networks. This is against our original purpose of using recursive networks, which is avoid introducing new parameters while stacking more layers. In addition, using different reconstruction nets no longer effectively regularizes the network. The second difference is that Lee et al. \cite{lee2014deeply} discards all intermediate classifiers during testing. However, an ensemble of all intermediate predictions significantly boosts the performance. The final output from the ensemble is also supervised.

Our recursive-supervision naturally eases the difficulty of training recursive networks. Backpropagation goes through a small number of layers if supervising signal goes directly from loss layer to early recursion. Summing all gradients backpropagated from different prediction losses gives a smoothing effect. The adversarial effect of vanishing/exploding gradients along one backpropagation path is alleviated.

Moreover, the importance of picking the optimal number of recursions is reduced as our supervision enables utilizing predictions from all intermediate layers. If recursions are too deep for the given task, we expect the weight for late predictions to be low while early predictions receive high weights.

By looking at weights of predictions, we can figure out the marginal gain from additional recursions. 

We present an expanded CNN structure of our model for illustration purposes in Figure \ref{fig:recursive_supervision}(c). If parameters are not allowed to be shared and CNN chains vary their depths, the number of free parameters grows fast (quadratically).

\textbf{Skip-Connection} Now we describe our second extension: skip-connection. For SR, input and output images are highly correlated. Carrying most if not all of input values until the end of the network is inevitable but very inefficient. Due to gradient problems, exactly learning a simple linear relation between input and output is very difficult if many recursions exist in between them.  

We add a layer skip \cite{bishop2006pattern} from input to the reconstruction net. Adding layer skips is successfully used for a semantic segmentation network \cite{long2014fully} and we employ a similar idea. Now input image is directly fed into the reconstruction net whenever it is used during recursions. Our skip-connection has two advantages. First, network capacity to store the input signal during recursions is saved. Second, the exact copy of input signal can be used during target prediction. 

Our skip-connection is simple yet very effective. In super-resolution, LR and HR images are vastly similar. In most regions, differences are zero and only small number of locations have non-zero values. For this reason, several super-resolution methods \cite{Timofte2013, Timofte, bevilacqua2012,bevilacqua2013super} predict image details only. Similarly, we find that this domain-specific knowledge significantly improves our learning procedure. 

\textbf{Mathematical Formulation} Each intermediate prediction under recursive-supervision (Figure \ref{fig:recursive_supervision}(a)) is 
\begin{equation}
\hat{{\bf y}}_{d} = f_3({\bf x}, g^{(d)}(f_1({\bf x}))),
\end{equation}
for $d=1,2,\dots,D$, where $f_3$ now takes two inputs, one from skip-connection. Reconstruction net with skip-connection can take various functional forms. For example, input can be concatenated to the feature maps $H_d$. As the input is an interpolated input image (roughly speaking, $\hat{\bf y} \approx {\bf x}$), we find $f_3({\bf x}, H_d) = {\bf x} + f_3(H_d)$ is enough for our purpose. More sophisticated functions for merging two inputs to $f_3$ will be explored in the future. 

Now, the final output is the weighted average of all intermediate predictions:
\begin{equation}
\hat{{\bf y}} = \sum_{d=1}^{D} w_d \cdot \hat{{\bf y}}_d.
\end{equation}
where $w_d$ denotes the weights of predictions reconstructed from each intermediate hidden state during recursion. These weights are learned during training.

\begin{table*}
\begin{center}
\setlength{\tabcolsep}{2pt}
\small
\begin{tabular}{ | c | c | c | c | c | c | c | c | }
\hline
\multirow{2}{*}{Dataset} & \multirow{2}{*}{Scale} & Bicubic & A+ \cite{Timofte} & SRCNN \cite{dong2014image} & RFL \cite{schulter2015fast} & SelfEx \cite{Huang-CVPR-2015} & DRCN (Ours)\\
 & & PSNR/SSIM & PSNR/SSIM & PSNR/SSIM & PSNR/SSIM & PSNR/SSIM & PSNR/SSIM\\
\hline
\hline
\multirow{3}{*}{Set5} & $\times$2 & 33.66/0.9299 & 36.54/{\color{blue}0.9544} & {\color{blue}36.66}/0.9542 & 36.54/0.9537 & 36.49/0.9537 & {\color{red}37.63}/{\color{red}0.9588}\\
 & $\times$3 & 30.39/0.8682 & 32.58/0.9088 & {\color{blue}32.75}/0.9090 & 32.43/0.9057 & 32.58/{\color{blue}0.9093} & {\color{red}33.82}/{\color{red}0.9226}\\
 & $\times$4 & 28.42/0.8104 & 30.28/0.8603 & {\color{blue}30.48}/{\color{blue}0.8628} & 30.14/0.8548 & 30.31/0.8619 & {\color{red}31.53}/{\color{red}0.8854}\\
\hline
\hline
\multirow{3}{*}{Set14} & $\times$2 & 30.24/0.8688 & 32.28/0.9056 & {\color{blue}32.42}/{\color{blue}0.9063} & 32.26/0.9040 & 32.22/0.9034 & {\color{red}33.04}/{\color{red}0.9118}\\
 & $\times$3 & 27.55/0.7742 & 29.13/0.8188 & {\color{blue}29.28}/{\color{blue}0.8209} & 29.05/0.8164 & 29.16/0.8196 & {\color{red}29.76}/{\color{red}0.8311}\\
 & $\times$4 & 26.00/0.7027 & 27.32/0.7491 & {\color{blue}27.49}/0.7503 & 27.24/0.7451 & 27.40/{\color{blue}0.7518} & {\color{red}28.02}/{\color{red}0.7670}\\
\hline
\hline
\multirow{3}{*}{B100} & $\times$2 & 29.56/0.8431 & 31.21/0.8863 & {\color{blue}31.36}/{\color{blue}0.8879} & 31.16/0.8840 & 31.18/0.8855 & {\color{red}31.85}/{\color{red}0.8942}\\
 & $\times$3 & 27.21/0.7385 & 28.29/0.7835 & {\color{blue}28.41}/{\color{blue}0.7863} & 28.22/0.7806 & 28.29/0.7840 & {\color{red}28.80}/{\color{red}0.7963}\\
 & $\times$4 & 25.96/0.6675 & 26.82/0.7087 & {\color{blue}26.90}/0.7101 & 26.75/0.7054 & 26.84/{\color{blue}0.7106} & {\color{red}27.23}/{\color{red}0.7233}\\
\hline
\hline
\multirow{3}{*}{Urban100} & $\times$2 & 26.88/0.8403 & 29.20/0.8938 & 29.50/0.8946 & 29.11/0.8904 & {\color{blue}29.54}/{\color{blue}0.8967} & {\color{red}30.75}/{\color{red}0.9133}\\
 & $\times$3 & 24.46/0.7349 & 26.03/0.7973 & 26.24/0.7989 & 25.86/0.7900 & {\color{blue}26.44}/{\color{blue}0.8088} & {\color{red}27.15}/{\color{red}0.8276}\\
 & $\times$4 & 23.14/0.6577 & 24.32/0.7183 & 24.52/0.7221 & 24.19/0.7096 & {\color{blue}24.79}/{\color{blue}0.7374} & {\color{red}25.14}/{\color{red}0.7510}\\
\hline
\end{tabular}
\caption{Benchmark results. Average PSNR/SSIMs for scale factor $\times$2, $\times$3 and $\times$4 on datasets Set5, Set14, B100 and Urban100. {\color{red}Red color} indicates the best performance and {\color{blue}blue color} refers the second best.}
\label{tbl:benchmark}
\end{center}
\end{table*}

\begin{figure*}
\begin{adjustwidth}{0.5cm}{0.5cm}
\begin{center}
\small
\setlength{\tabcolsep}{3pt}
\begin{tabular}{  c  c  c  c  c  c  }
{\graphicspath{{figs/figDRCN/}}\includegraphics[width=0.15\textwidth]{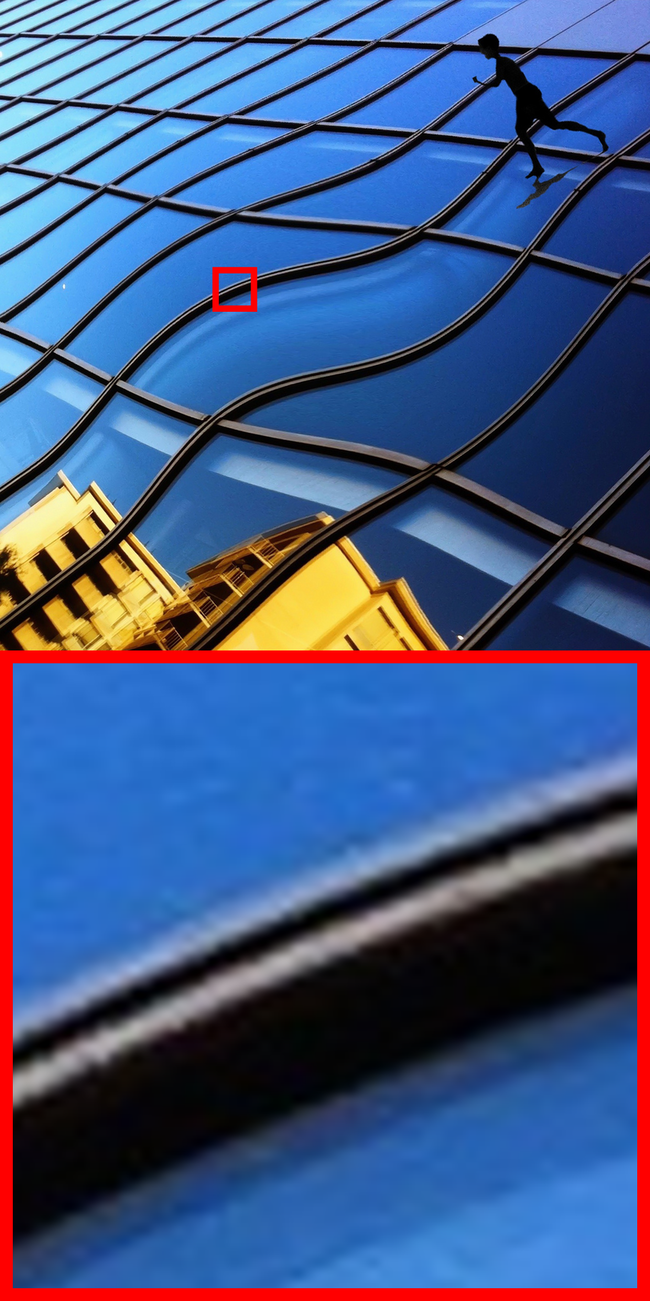}}
& {\graphicspath{{figs/figDRCN/}}\includegraphics[width=0.15\textwidth]{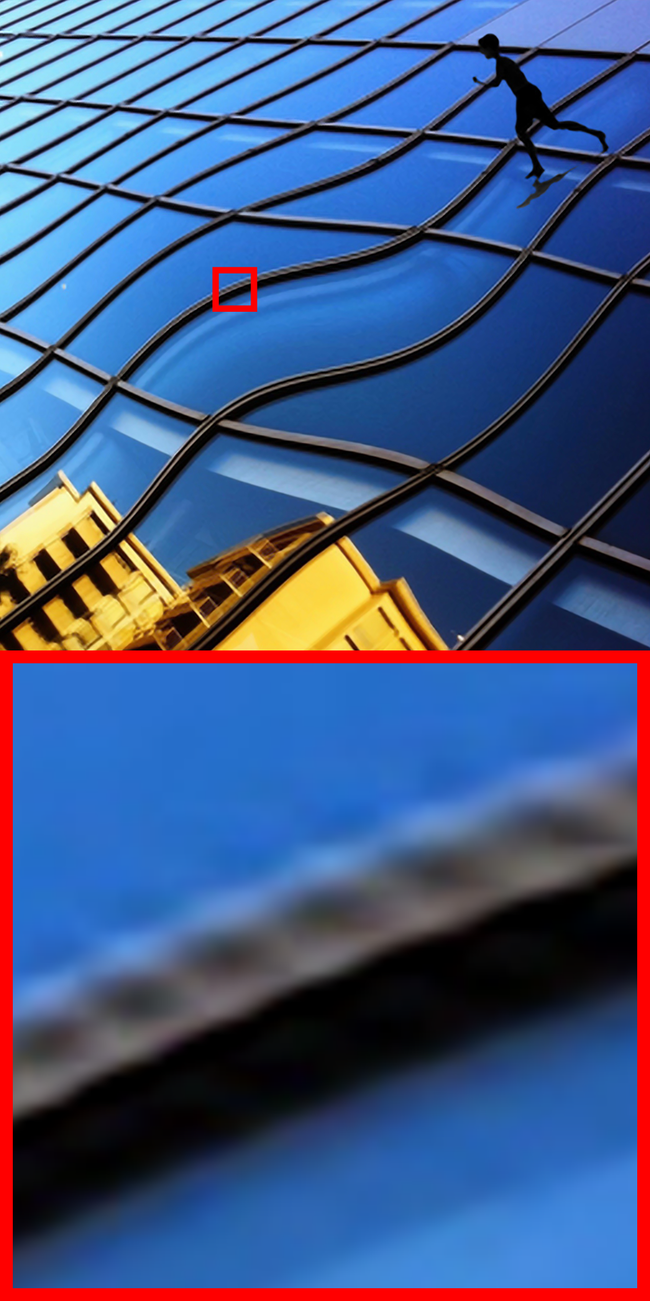}}
& {\graphicspath{{figs/figDRCN/}}\includegraphics[width=0.15\textwidth]{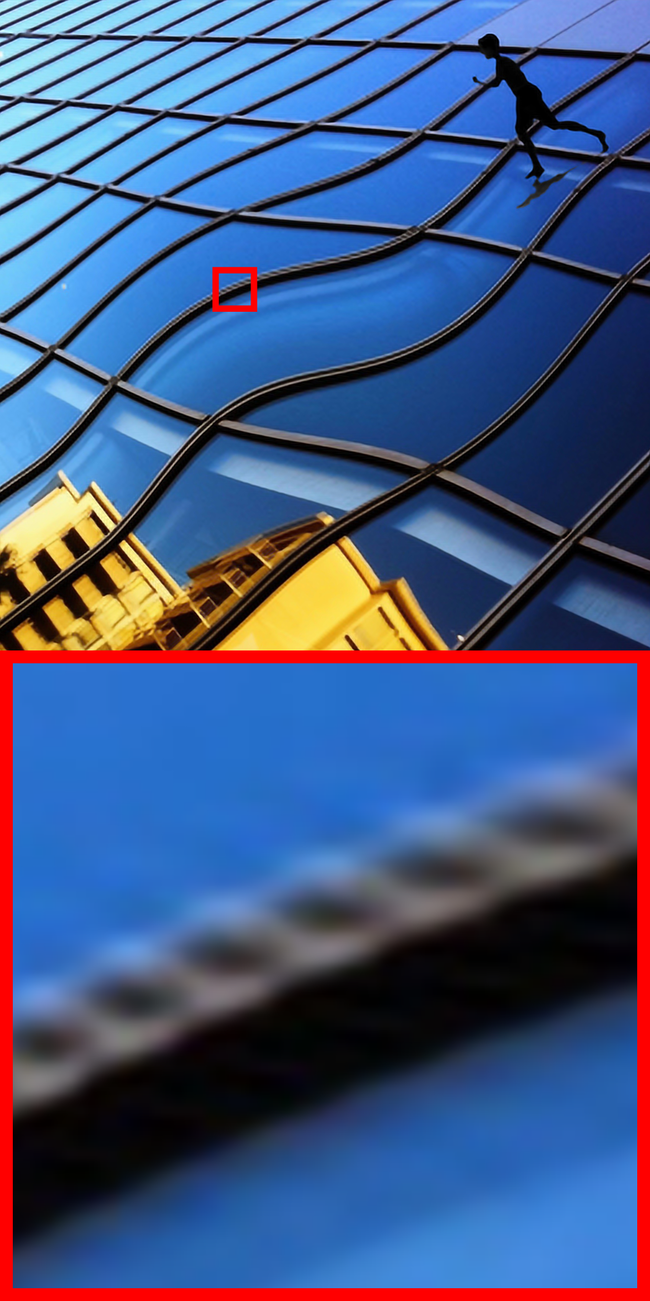}}
& {\graphicspath{{figs/figDRCN/}}\includegraphics[width=0.15\textwidth]{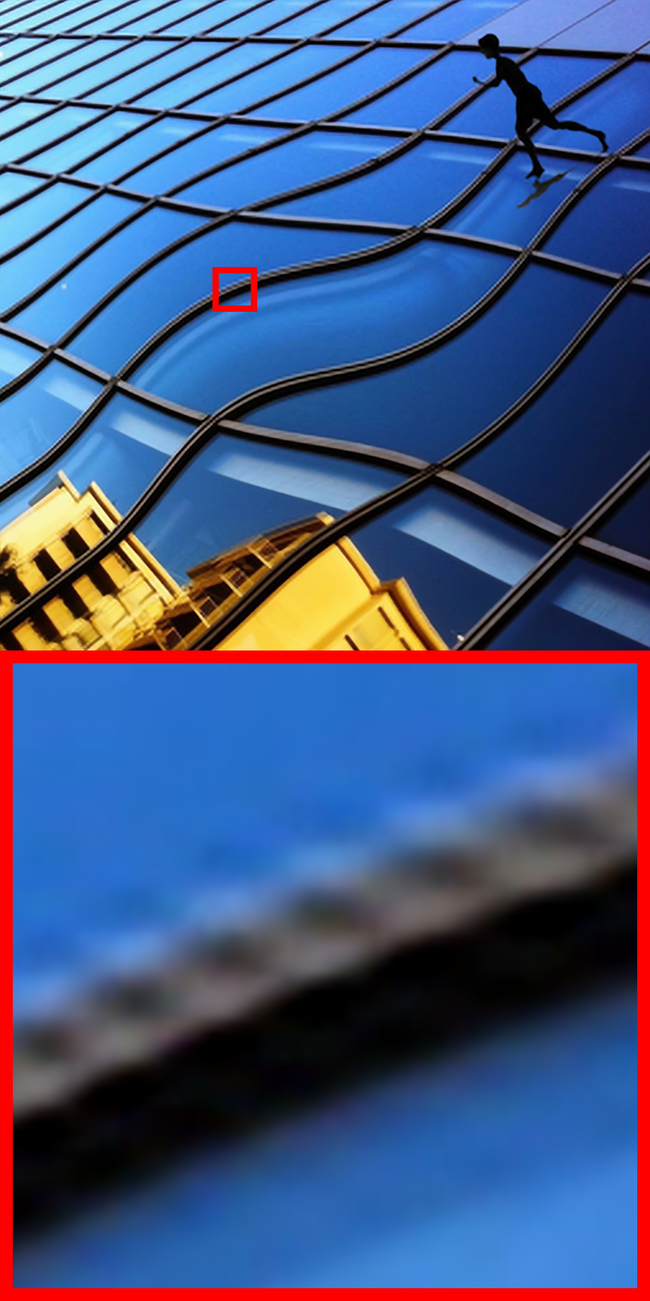}}
& {\graphicspath{{figs/figDRCN/}}\includegraphics[width=0.15\textwidth]{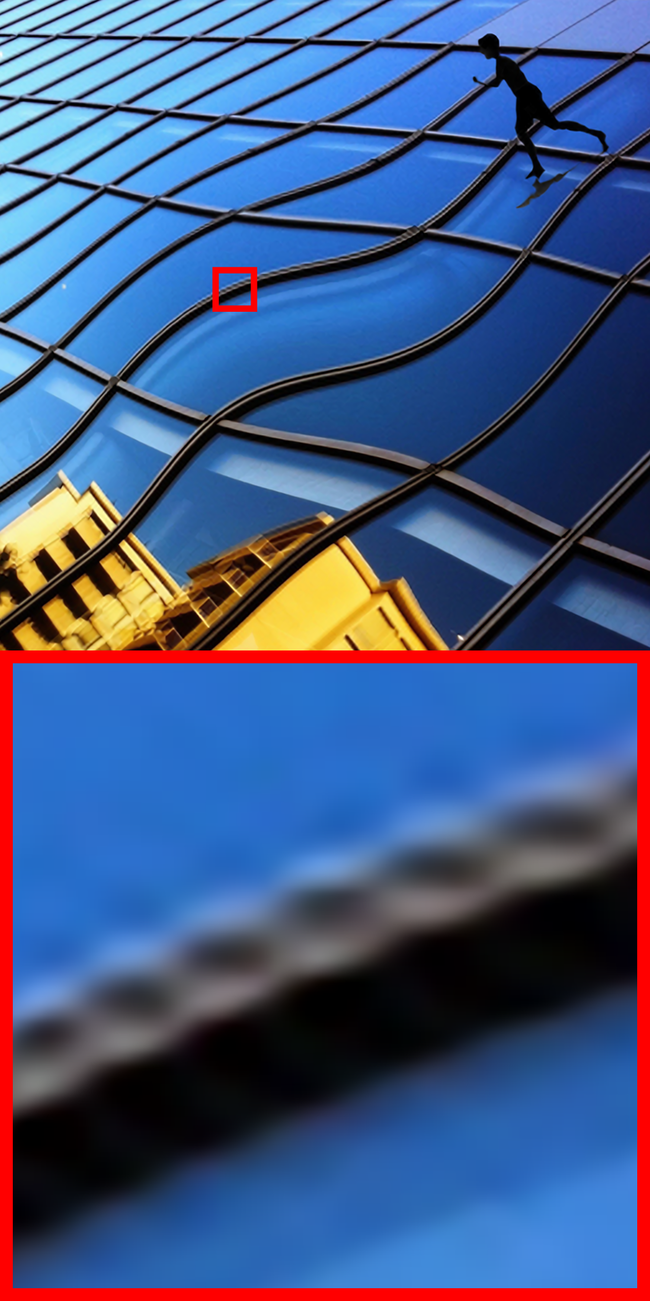}}
& {\graphicspath{{figs/figDRCN/}}\includegraphics[width=0.15\textwidth]{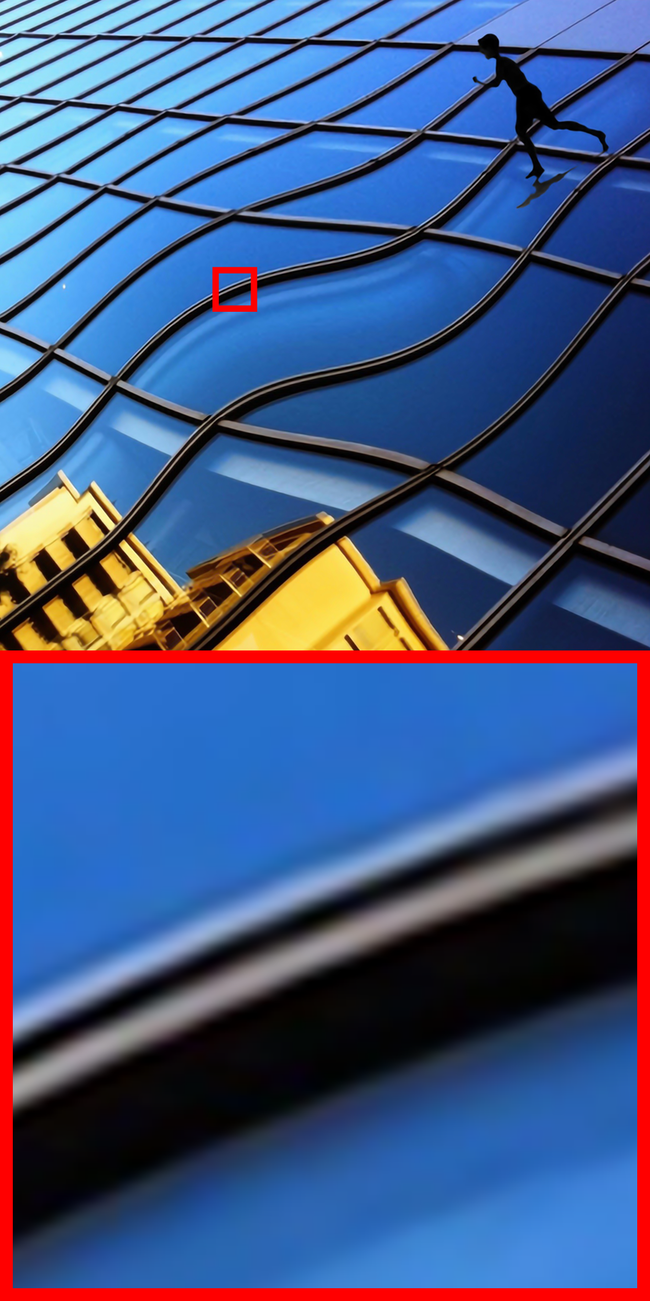}}
\\
Ground Truth& A+ \cite{Timofte}& SRCNN \cite{dong2014image}& RFL \cite{schulter2015fast}& SelfEx \cite{Huang-CVPR-2015}& DRCN (Ours)\\
(PSNR, SSIM)& (29.83, 0.9102)& (29.97, 0.9092)& (29.61, 0.9026)& ({\color{blue}{30.73}}, {\color{blue}{0.9193}})& ({\color{red}{32.17}}, {\color{red}{0.9350}})\\
\end{tabular}
\caption{Super-resolution results of ``img082"(\textit{Urban100}) with scale factor $\times$4. Line is straightened and sharpened in our result, whereas other methods give blurry lines. Our result seems visually pleasing.}
\label{fig:img1}
\end{center}
\end{adjustwidth}
\end{figure*}

\begin{figure*}
\begin{adjustwidth}{0.5cm}{0.5cm}
\begin{center}
\small
\setlength{\tabcolsep}{3pt}
\begin{tabular}{  c  c  c  c  c  c  }
{\graphicspath{{figs/figDRCN/}}\includegraphics[width=0.15\textwidth]{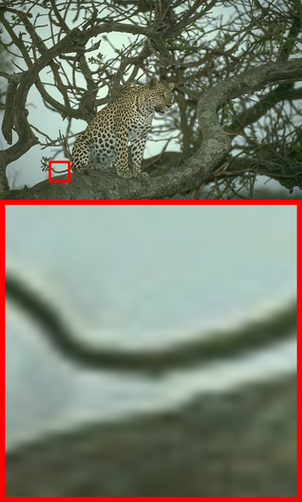}}
& {\graphicspath{{figs/figDRCN/}}\includegraphics[width=0.15\textwidth]{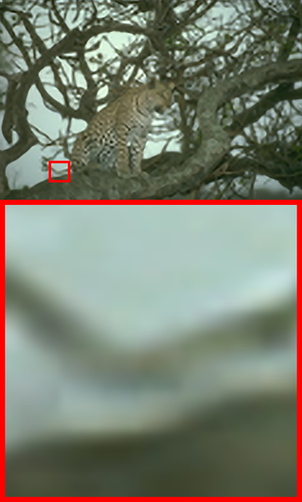}}
& {\graphicspath{{figs/figDRCN/}}\includegraphics[width=0.15\textwidth]{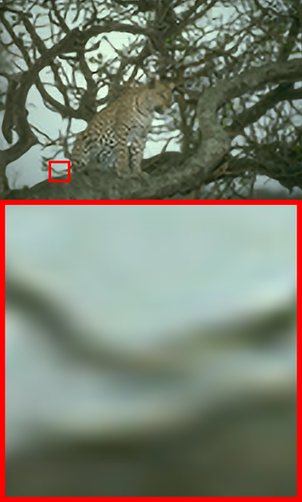}}
& {\graphicspath{{figs/figDRCN/}}\includegraphics[width=0.15\textwidth]{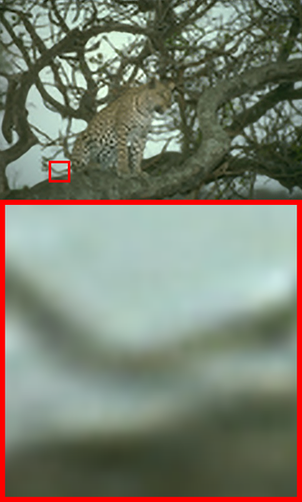}}
& {\graphicspath{{figs/figDRCN/}}\includegraphics[width=0.15\textwidth]{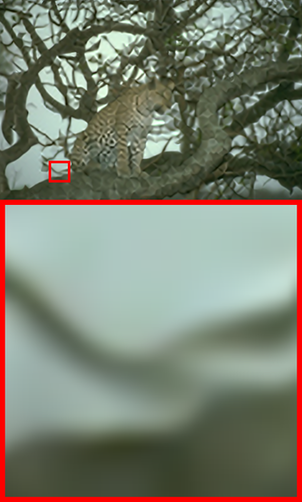}}
& {\graphicspath{{figs/figDRCN/}}\includegraphics[width=0.15\textwidth]{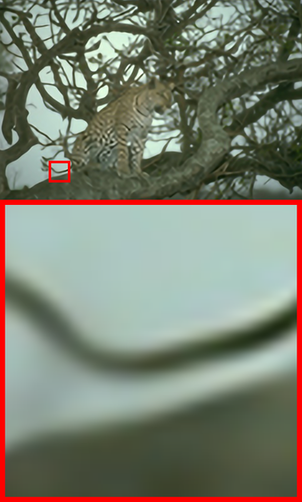}}
\\
Ground Truth& A+ \cite{Timofte}& SRCNN \cite{dong2014image}& RFL \cite{schulter2015fast}& SelfEx \cite{Huang-CVPR-2015}& DRCN (Ours)\\
(PSNR, SSIM)& (23.53, 0.6977)& ({\color{blue}{23.79}}, {\color{blue}{0.7087}})& (23.53, 0.6943)& (23.52, 0.7006)& ({\color{red}{24.36}}, {\color{red}{0.7399}})\\
\end{tabular}
\caption{Super-resolution results of ``134035" (\textit{B100}) with scale factor $\times$4. Our result shows a clear separation between branches while in other methods, branches are not well separated.  }
\label{fig:img2}
\end{center}
\end{adjustwidth}
\end{figure*}

\begin{figure*}
\begin{adjustwidth}{0.5cm}{0.5cm}
\begin{center}
\small
\setlength{\tabcolsep}{3pt}
\begin{tabular}{  c  c  c  c  c  c  }
{\graphicspath{{figs/figDRCN/}}\includegraphics[width=0.15\textwidth]{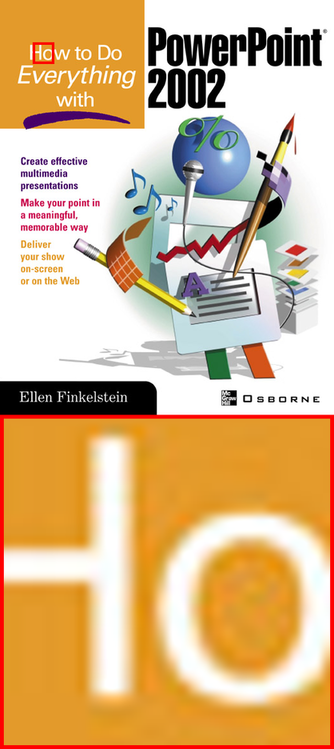}}
& {\graphicspath{{figs/figDRCN/}}\includegraphics[width=0.15\textwidth]{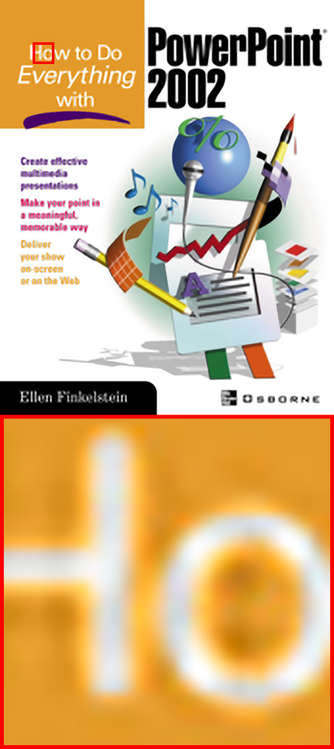}}
& {\graphicspath{{figs/figDRCN/}}\includegraphics[width=0.15\textwidth]{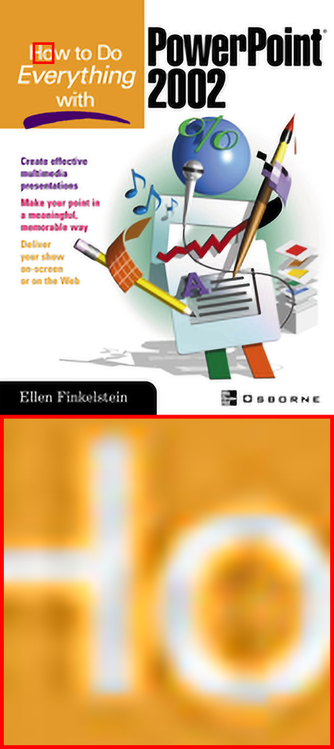}}
& {\graphicspath{{figs/figDRCN/}}\includegraphics[width=0.15\textwidth]{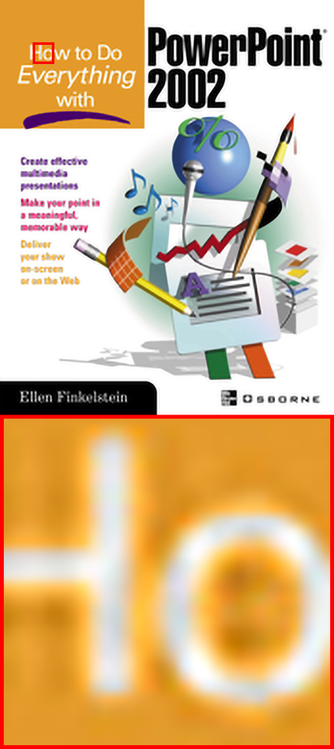}}
& {\graphicspath{{figs/figDRCN/}}\includegraphics[width=0.15\textwidth]{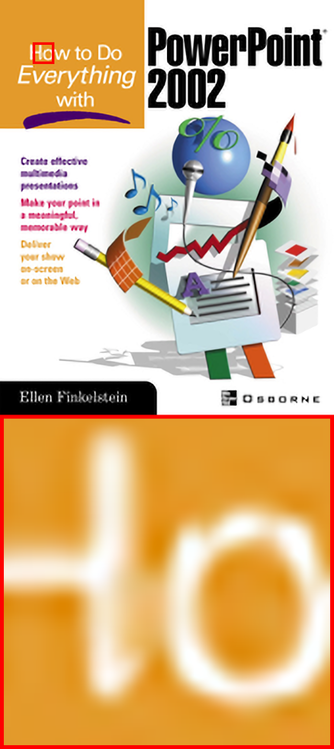}}
& {\graphicspath{{figs/figDRCN/}}\includegraphics[width=0.15\textwidth]{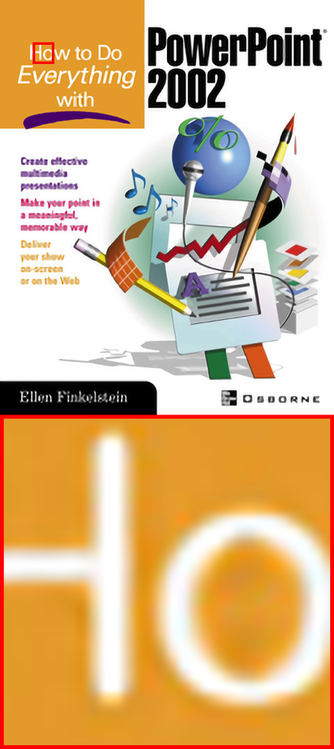}}
\\
Ground Truth& A+ \cite{Timofte}& SRCNN \cite{dong2014image}& RFL \cite{schulter2015fast}& SelfEx \cite{Huang-CVPR-2015}& DRCN (Ours)\\
(PSNR, SSIM)& (26.09, 0.9342)& (27.01, 0.9365)& (25.91, 0.9254)& ({\color{blue}{27.10}}, {\color{blue}{0.9483}})& ({\color{red}{27.66}}, {\color{red}{0.9608}})\\
\end{tabular}
\caption{Super-resolution results of ``ppt3" (\textit{Set14}) with scale factor $\times$3. Texts in DRCN are sharp while, in other methods, character edges are blurry.}
\label{fig:img3}
\end{center}
\end{adjustwidth}
\end{figure*}

\begin{figure*}
\begin{adjustwidth}{0.5cm}{0.5cm}
\begin{center}
\small
\setlength{\tabcolsep}{3pt}
\begin{tabular}{  c  c  c  c  c  c  }
{\graphicspath{{figs/figDRCN/}}\includegraphics[width=0.15\textwidth]{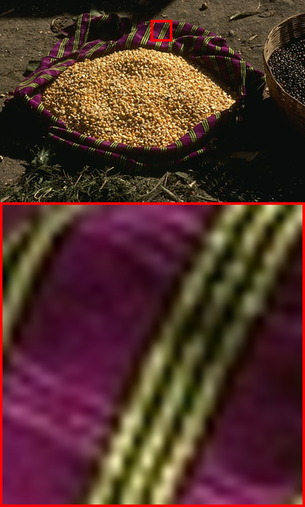}}
& {\graphicspath{{figs/figDRCN/}}\includegraphics[width=0.15\textwidth]{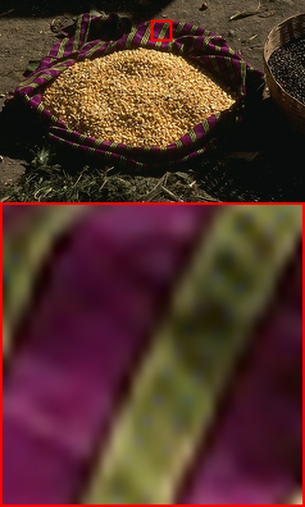}}
& {\graphicspath{{figs/figDRCN/}}\includegraphics[width=0.15\textwidth]{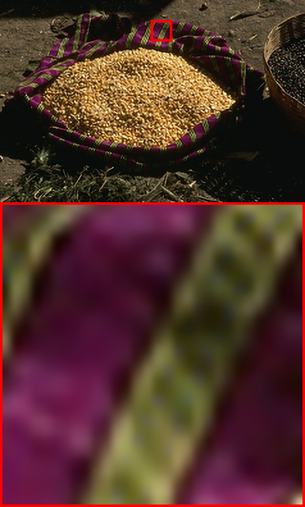}}
& {\graphicspath{{figs/figDRCN/}}\includegraphics[width=0.15\textwidth]{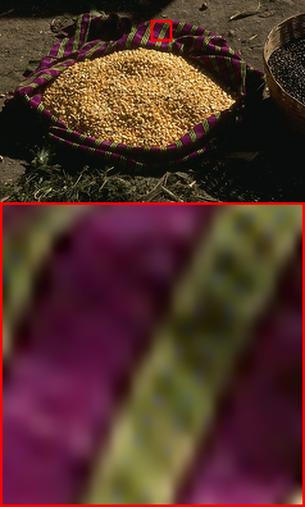}}
& {\graphicspath{{figs/figDRCN/}}\includegraphics[width=0.15\textwidth]{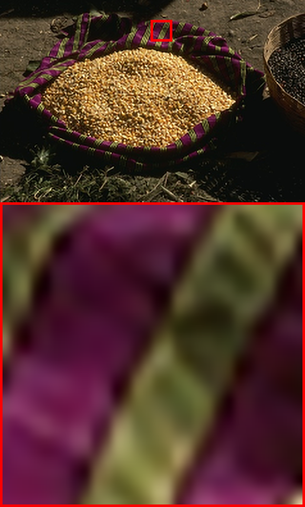}}
& {\graphicspath{{figs/figDRCN/}}\includegraphics[width=0.15\textwidth]{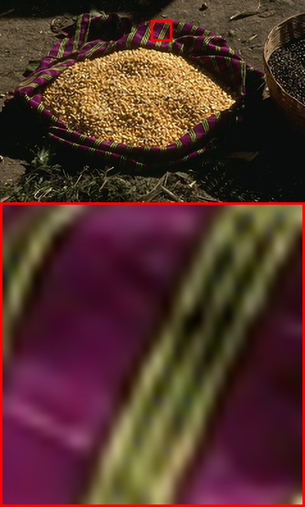}}
\\
Ground Truth& A+ \cite{Timofte}& SRCNN \cite{dong2014image}& RFL \cite{schulter2015fast}& SelfEx \cite{Huang-CVPR-2015}& DRCN (Ours)\\
(PSNR, SSIM)& (24.24, 0.8176)& ({\color{blue}{24.48}}, {\color{blue}{0.8267}})& (24.24, 0.8137)& (24.16, 0.8145)& ({\color{red}{24.76}}, {\color{red}{0.8385}})\\
\end{tabular}
\caption{Super-resolution results of ``58060" (\textit{B100}) with scale factor $\times$2. A three-line stripe in ground truth is also observed in DRCN, whereas it is not clearly seen in results of other methods.}
\label{fig:img4}
\end{center}
\end{adjustwidth}
\end{figure*}


\subsection{Training}

\textbf{Objective} We now describe the training objective used to find optimal parameters of our model. Given a training dataset $\{{\bf x}^{(i)},{\bf y}^{(i)}\}{}_{i=1}^{N}$, our goal is to find the best model $f$ that accurately predicts values $\mathbf{\hat{y}}=f(\mathbf{x})$.

In the least-squares regression setting, typical in SR, the mean squared error $\frac{1}{2}||\mathbf{y}-f(\mathbf{x})||^{2}$
averaged over the training set is minimized. This favors high Peak Signal-to-Noise
Ratio (PSNR), a widely-used evaluation criteria. 

With recursive-supervision, we have $D+1$ objectives to minimize: supervising $D$ outputs from recursions and the final output. For intermediate outputs, we have the loss function 
\begin{equation}
l_1(\theta) = \sum_{d=1}^D \sum_{i=1}^N \frac{1}{2DN}||{\bf y}^{(i)} -  \hat{\bf y}_d^{(i)} ||^{2},
\end{equation}
where $\theta$ denotes the parameter set and $\hat{\bf y}_d^{(i)}$ is the output from the $d$-th recursion. For the final output, we have 
\begin{equation}
l_2(\theta) = \sum_{i=1}^N \frac{1}{2N}||{\bf y}^{(i)} -  \sum_{d=1}^D  w_d \cdot \hat{\bf y}_d^{(i)} ||^{2}
\end{equation}

Now we give the final loss function $L(\theta)$. The training is regularized by weight decay ($L_2$ penalty multiplied by $\beta$). 
\begin{equation}
L(\theta)  =\alpha  l_1(\theta) + (1 - \alpha) l_2(\theta) + \beta ||\theta||^2,
\end{equation}
where $\alpha$ denotes the importance of the companion objective on the intermediate outputs and $\beta$ denotes the multiplier of weight decay.   Setting $\alpha$ high makes the training procedure stable as early recursions easily converge. As training progresses, $\alpha$ decays to boost the performance of the final output. 

Training is carried out by optimizing the regression objective using mini-batch gradient descent based on back-propagation (LeCun et al. \cite{lecun1998gradient}). We implement our model using the \textit{MatConvNet}\footnote{\url{ http://www.vlfeat.org/matconvnet/}} package \cite{arXiv:1412.4564}.

\section{Experimental Results}
In this section, we evaluate the performance of our method on several datasets. We first describe datasets used for training and testing our method. Next, our training setup is given. We give several experiments for understanding our model properties. The effect of increasing the number of recursions is investigated. Finally, we compare our method with several state-of-the-art methods. 

\subsection{Datasets}
For training, we use 91 images proposed in Yang et al. \cite{yang2010image} for all experiments. For testing, we use four datasets. Datasets \textit{Set5} \cite{bevilacqua2012} and \textit{Set14} \cite{zeyde2012single} are often used for benchmark \cite{Timofte,Timofte2013,dong2014image}. Dataset \textit{B100}  consists of natural images in the Berkeley Segmentation Dataset \cite{Martin2001}. Finally, dataset \textit{Urban100}, urban images recently provided by Huang et al. \cite{Huang-CVPR-2015}, is very interesting as it contains many challenging images failed by existing methods.

\begin{figure}
\begin{adjustwidth}{0cm}{-0.0cm}
\centering
{\graphicspath{{figs/graph1/}}\includegraphics[width=0.4\textwidth]{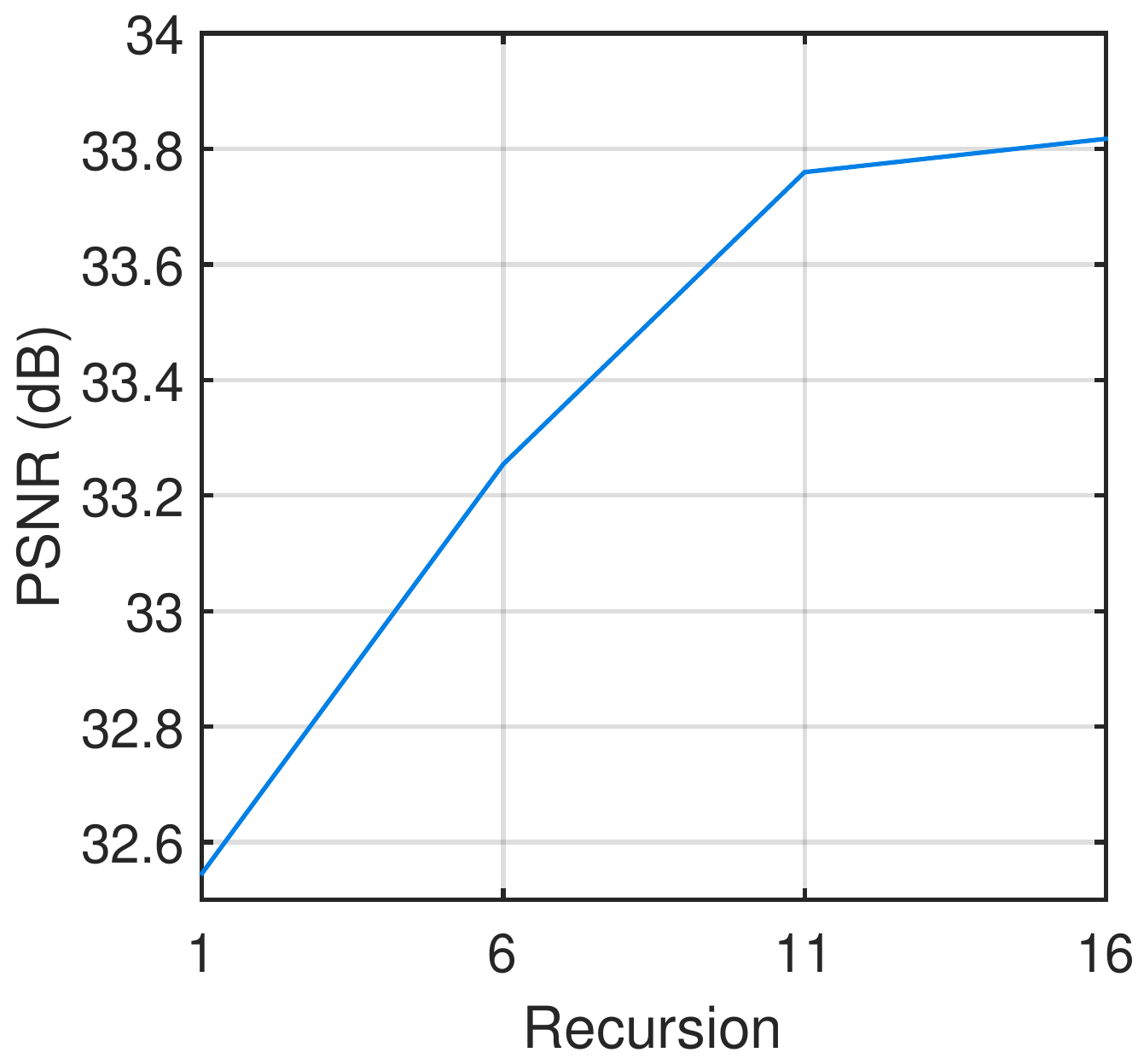}}
\caption{Recursion versus Performance for the scale factor $\times$3 on the dataset \textit{Set5}. More recursions yielding larger receptive fields lead to better performances.}\end{adjustwidth}
\label{fig:more}
\end{figure}

\begin{figure}
\centering
{\graphicspath{{figs/graph1/}}\includegraphics[width=0.45\textwidth]{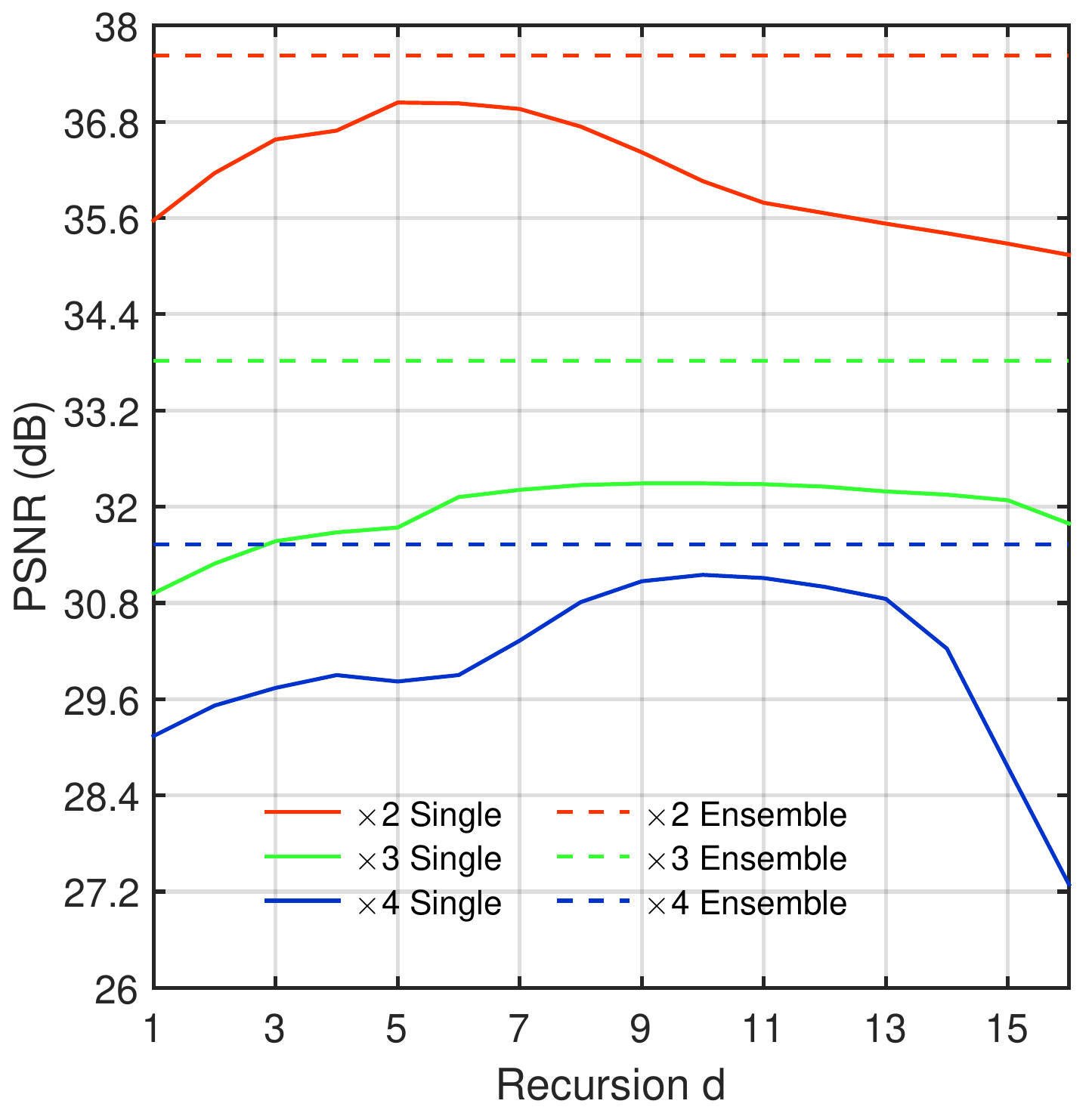}}
\caption{Ensemble effect. Prediction made from intermediate recursions are evaluated. There is no single recursion depth that works the best across all scale factors. Ensemble of intermediate predictions significantly improves performance. }
\label{fig:ensemble}
\end{figure}

\subsection{Training Setup}
%
%
%
%
%
%

We use 16 recursions unless stated otherwise. When unfolded, the longest chain from the input to the output passes 20 conv. layers (receptive field of 41 by 41). We set the momentum parameter to 0.9 and weight decay to 0.0001. We use 256 filters of the size $3 \times 3$ for all weight layers. Training images are split into 41 by 41 patches with stride 21 and 64 patches are used as a mini-batch for stochastic gradient descent. 

For initializing weights in non-recursive layers, we use the method described in He et al. \cite{he2015delving}. For recursive convolutions, we set all weights to zero except self-connections (connection to the same neuron in the next layer) \cite{socher2012semantic, le2015simple}.  Biases are set to zero.

Learning rate is initially set to 0.01 and then decreased by a factor of 10 if the validation error does not decrease for 5 epochs. If learning rate is less than $10^{-6}$, the procedure is terminated. Training roughly takes 6 days on a machine using one Titan X GPU. 

\subsection{Study of Deep Recursions}
We study the effect of increasing recursion depth. We trained four models with different numbers of recursions: 1, 6, 11, and 16. Four models use the same number of parameters except the weights used for ensemble. In Figure {\color{red}8}, it is shown that as more recursions are performed, PSNR measures increase. Increasing recursion depth with a larger image context and more nonlinearities boosts performance. The effect of ensemble is also investigated. We first evaluate intermediate predictions made from recursions (Figure \ref{fig:ensemble}). The ensemble output significantly improves performances of individual predictions. 

\subsection{Comparisons with State-of-the-Art Methods}
We provide quantitative and qualitative comparisons. For benchmark, we use public code for A+ \cite{Timofte}, SRCNN \cite{dong2014image}, RFL \cite{schulter2015fast} and  SelfEx \cite{Huang-CVPR-2015}. We deal with luminance components only as similarly done in other methods because human vision is much more sensitive to details in intensity than in color.

As  some methods such as A+ \cite{Timofte} and  RFL \cite{schulter2015fast} do not predict image boundary, they require cropping pixels near borders. For our method, this procedure is unnecessary as our network predicts the full-sized image. For fair comparison, however, we also crop pixels to the same amount. PSNRs can be slightly different from original papers as existing methods use slightly different evaluation frameworks. We use the public evaluation code used in \cite{Huang-CVPR-2015}.

In Table \ref{tbl:benchmark}, we provide a summary of quantitative evaluation on several datasets. 
Our method outperforms all existing methods in all datasets and scale factors (both PSNR and SSIM). In Figures \ref{fig:img1}, \ref{fig:img2}, \ref{fig:img3} and \ref{fig:img4}, example images are given. Our method produces relatively sharp edges respective to patterns. In contrast, edges in other images are blurred. Our method takes a second to process a $288 \times 288$ image on a GPU Titan X.

\section{Conclusion}
In this work, we have presented a super-resolution method using a deeply-recursive convolutional network. Our network efficiently reuses weight parameters while exploiting a large image context. To ease the difficulty of training the model, we use recursive-supervision and skip-connection. We have demonstrated that our method outperforms existing methods by a large margin on benchmarked images. In the future, one can try more recursions in order to use image-level context. We believe our approach is readily applicable to other image restoration problems such as denoising and compression artifact removal.

{\small
	\bibliographystyle{ieee}
	\bibliography{DRCN}
}
\end{document}